\documentclass[review, sort&compress]{arxiv}

\usepackage{natbib}
\usepackage{array}
\usepackage{url}

\usepackage[breaklinks]{hyperref}
\usepackage{breakurl}

\usepackage{amssymb,amsmath,subcaption}
\usepackage{graphicx}\graphicspath{{figs/}{eps/}}
\usepackage{psfrag}
\usepackage{ulem}
\usepackage{algorithmic}
\usepackage{algorithm}
\usepackage{varioref}
\usepackage{booktabs}
\usepackage{tabularx}
\journal{ArXiV}

\newcommand{\mat}[1]{\left[\begin{array}{#1}}

\usepackage{url}
\begin{document}

\begin{frontmatter}

\title{Feature Selection through Minimization of the VC dimension\footnote{For commercial use or licensing of the MCM and its variants, contact the Foundation for Innovation and Technology Transfer, IIT Delhi.}}
\address[ee]{Department of Electrical Engineering}
\address[cse]{Department of Computer Science \& Engineering}


\author[ee]{Jayadeva,~\textit{Senior Member,~IEEE}}
\ead{jayadeva@ee.iitd.ac.in}
\author[cse]{Sanjit S. Batra}
\author[ee]{Siddharth Sabharwal}
\address{Indian Institute of Technology, Delhi}

\begin{abstract}
Feature selection involes identifying the most relevant subset of input features, with a view to improving generalization of predictive models by reducing overfitting. Directly searching for the most relevant combination of attributes is NP-hard. Variable selection is of critical importance in many applications, such as micro-array data analysis, where selecting a small number of discriminative features is crucial to developing useful models of disease mechanisms, as well as for prioritizing targets for drug discovery. In this paper, we use very new results in machine learning to develop a novel feature selection strategy. The recently proposed Minimal Complexity Machine (MCM) provides a way to learn a hyperplane classifier by minimizing an exact (\boldmath{$\Theta$}) bound on its VC dimension. It is well known that a lower VC dimension contributes to good generalization. Experimental results show that the MCM learns very sparse representations; on many datasets, the kernel MCM yields comparable or better test set accuracies while using less than one-tenth the number of support vectors. For a linear hyperplane classifier in the input space, the VC dimension is upper bounded by the number of features; hence, a linear classifier with a small VC dimension is parsimonious in the set of features it employs. In this paper, we use the linear MCM to learn a classifier in which a large number of weights are zero; features with non-zero weights are the ones that are chosen. Selected features are used to learn a kernel SVM classifier. On a number of benchmark datasets, the features chosen by the linear MCM yield comparable or better test set accuracy than when methods such as ReliefF and FCBF are used for the task. The linear MCM typically chooses $\frac{1}{10}$-th the number of attributes chosen by the other methods; on some very high dimensional datasets, the proposed approach chooses about $0.6\%$ of the features; in comparison, ReliefF and FCBF choose 70 to 140 \textit{times} more features, thus demonstrating that minimizing the VC dimension may provide a new, and very effective route for feature selection and for learning sparse representations. 
\end{abstract}


\end{frontmatter}

\section{Introduction}
Feature selection is an important focal point in machine learning, and is also termed as variable selection, variable subset selection, or attribute selection. In many machine learning scenarios such as text analytics and gene micro-array data analysis, data samples have hundreds to thousands of attributes. Irrelevant or redundant features negatively impact learning methods, by introducing noise, contributing to overfitting, and leading to poorer generalization. Feature selection involves identifying the most relevant subset of input features for use in model construction. The goals of attribute selection include improving generalization of predictive models by reducing overfitting, decreasing training times, and facilitating the construction of more interpretable models. While feature selection can be applied to both supervised and unsupervised learning. Recently, there has been rising research interest into the field of feature selection for unsupervised learning \cite{varshavsky2007unsupervised} and into feature selection for supervised regression \cite{navot2005nearest}. 

In the contex of supervised classification, feature selection techniques can be organized into three categories, depending on how they combine the feature selection with the construction of the classification model, namely, filter methods, wrapper methods and embedded methods. A recent review of feature selection methods in bio-informatics may be found in \cite{saeys2007review}. A less recent collection of many diverse approaches includes \cite{Bi20031229, Perkins20031333, Bekkerman20031183, Dhillon20031265, Rivals20031383, Stoppiglia20031399, Weston20031439, Bengio20031209, Reunanen20031371, Rakotomamonjy20031357, Torkkola20031415, Caruana20031245, Globerson20031307, Forman20031289, guyon2003introduction}. Bayesian approaches built around Bayesian PCA tackle the problem by incorporating sparsity inducing priors \cite{o2009review, bishop1999bayesian, nakajima2011bayesian, li2012preserving, bickel2010hierarchical, bishop1999variational}. 

\begin{figure}[hbtp]
        \centering	
                \includegraphics[scale=0.5]{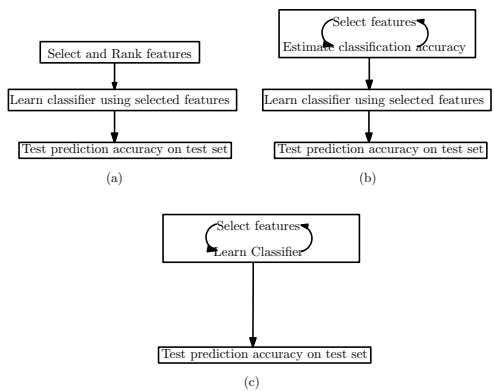}
                \caption{Different approaches to feature selection primarily differ in terms of the interaction between the feature selection process and the construction of the classifier that uses the features. Attribute selection methods may be broadly classified into (a) filter, (b) wrapper, and (c) embedded techniques.}
\end{figure}\label{fig1}

\textit{Filter methods}(Figure \ref{fig1}(a)) do not optimize for classification accuracy of the classifier directly, but attempt to select features by using measures such as the  ${\chi}^2$ statistic or the t-statistic. These methods are fast and hence, scaleable, but ignore feature dependencies and interaction of the feature selection search with the construction of the classifier. Some of the most influential filter techniques are the (\cite{yu2003feature},\cite{kira1992feature}). Different from filter techniques are \textit{wrapper methods}(Figure \ref{fig1}(b)). These methods embed the construction of the classifier within the feature selection process. They iterate over a defined feature subset space, and select the feature subset based on its power to improve the sample classification accuracy. Classical wrapper algorithms include forward selection and backward elimination(\cite{kohavi1997wrappers}). These methods have the ability to take into account feature dependencies, and there is an inherent interaction between the classifier and feature selection. However, wrapper methods face a higher risk of overfitting than filter methods and are computationally more expensive. \textit{Embedded methods}(Figure \ref{fig1}(c))combine classifier construction with feature selection, and hence feature selection is a by-product of the classification process. Embedded methods, such as the ones in (\cite{guyon2003introduction}), are computationally less expensive and are less prone to overfitting. However, they are specific to a learning machine.

Each of these three approaches has its advantages and disadvantages, the predominant distinguishing factors being computational speed and the risk of overfitting. A primary advantage of filter methods is that they are amenable to a theoretical analysis of their design, and this advantage can be leveraged. Recent information-theoretic feature selection methods(\cite{brown2012conditional}) attempt to use this advantage in order to derive feature relevance indices based on an objective function, instead of defining them based on heuristics.

Note that in each case, while the optimization may be done within a loop or in a feedforward manner, any optimization based approach would need a measure of the discriminative power of a feature or of a set of features. Ideally, one would like to employ a universal measure. The generalization ability of a learning machine may be measured by its Vapnik-Chervonenkis (VC) dimension. The VC dimension can be used to estimate a probabilistic upper bound on the test set error of a classifier. Vapnik showed, that with probability $1-\eta$, the following bound holds:
\begin{gather}
 \text{test set error} \leq \text{training set error} + \sqrt{\frac{\gamma(1 + \frac{log(2M)}{\gamma})-log(\frac{\eta}{4})}{M}}
\end{gather}

A small VC dimension leads to good generalization and low error rates on test data. This paper suggests a new approach to feature selection by asking a fundamental question: can we use a measure of the generalization ability to select features ? Such an approach, if feasible, would avoid overfitting, and would not be affected by the choice of classifier used to learn with the selected features. We show that this is indeed possible, and that feature selection can be achieved by learning a classifier through minimizing an exact bound on the VC dimension. The notion of an exact bound means that the objective being minimized bounds the VC dimension from both above and below; this means that the two are close to each other. The theory that allows us to do so is motivated by the recently proposed Minimal Complexity Machine (MCM) \cite{mcmneucom, mcmarxiv}. The MCM shows that it is possible to learn a hyperplane classifier by minimizing an exact bound on the VC dimension.

The number of features is also an upper bound on the VC dimension of a hyperplane classifier, and consequently, a small VC dimension hyperplane classifier also uses a small subset of discriminative features. Since these features also control the generalization error, they should form a good choice regardless of the classifier used. We demonstrate that this is indeed the case, by using the selected features to train a SVM classifier with a radial basis function (RBF) kernel. On selected competition benchmark datasets, the MCM typically chooses $\frac{1}{10}$-th the number of attributes chosen by the other methods; on some very high dimensional datasets, the proposed approach chooses about $0.6\%$ of the features; in comparison, ReliefF \cite{relieff} and FCBF \cite{fcbf} choose 70 to 140 \textit{times} more features. 
Yet, the test set accuracy, measured using a five fold cross validation methodology, is generally better when features selected by the MCM are used to learn the classifier.

The use of a SVM classifier on features selected by the MCM shows that a feature selection scheme that is directly driven by generalization error can find the best of the wrapper, embedded, and filter worlds. The MCM requires only a linear programming problem to be solved, and is therefore computationally attractive and scaleable to large datasets. The approach in this paper may be summarized as follows: we first train a linear MCM on a given high dimensional dataset. Because the MCM directly minimizes an exact bound on the VC dimension, the linear MCM classifier that is learnt is parsimonious in its use of features. The approach is inherently a multivariate filter, which does not ignore feature dependencies; this is now widely believed to be an important requirement for a successful filter method. Still, unlike classical filter methods which disregard the interaction between feature selection and learning the classifier, the MCM based approach unifies the two. This is because minimizing the VC dimension directly addresses generalization; it also means that learning with the chosen features will lead to better generalization regardless of the classifier used. A classifier with low VC dimension generalizes better, and the objective of minimizing the VC dimension for a linear MCM achieves feature selection. Once the relevant feature set has been identified, a kernel SVM classifier is learnt using the chosen attributes.

The remainder of this paper is organized as follows. Section \ref{linmcm} briefly describes the Minimal Complexity Machine (MCM) classifier. Section \ref{experimental} is devoted to a discussion of results obtained on selected benchmark datasets. Section \ref{conclusion} contains concluding remarks.

\section{The Linear Minimal Complexity Machine} \label{linmcm}

Consider a binary classification dataset with $n$-dimensional samples $x^i, i = 1, 2, ..., M$, where each sample is associated with a label $y_i \in \{+1, -1\}$. We assume that the dimension of the input samples is $n$, i.e. $x^i = (x_1^i, x_2^i, ..., x_n^i)^T$. The motivation for the MCM originates from some outstanding work on generalization \cite{shawe1996framework, shawetaylor98, vapnik98, scholkopf2002learning}. Vapnik \cite{vapnik98} showed that the VC dimension $\gamma$ for fat margin hyperplane classifiers with margin $d \geq d_{min}$ satisfies
\begin{equation}\label{eqnh}
\gamma \leq 1 + \operatorname{Min}(\frac{R^2}{d^2_{min}}, n)
\end{equation}
where $R$ denotes the radius of the smallest sphere enclosing all the training samples. Burges, in \cite{burges1998}, stated that \textit{``the above arguments strongly suggest that algorithms that minimize $\frac{R^2}{d^2}$ can be expected to give better generalization performance. Further evidence for this is found in the following theorem of (Vapnik, 1998), which we quote without proof''}.\\

Consider the case of a linearly separable dataset. By definition, there exists a hyperplane that can classify these points with zero error. Let the separating hyperplane be given by
\begin{equation}
 u^Tx + v = 0.
\end{equation}

Let us denote
\begin{gather}
 h = \frac{\operatorname*{Max}_{i = 1, 2, ..., M} \;y_i(u^T x^i + v)}{\operatorname*{Min}_{i = 1, 2, ..., M} \;y_i(u^T x^i + v)}.
\end{gather}
In \cite{mcmneucom, mcmarxiv}, we show that there exist constants $\alpha, \beta > 0$, $\alpha, \beta \in \Re$ such that
\begin{equation}\label{exactbound}
 \alpha h^2 \leq \gamma \leq \beta h^2,
\end{equation}
or, in other words, $h^2$ constitutes a tight or exact ($\theta$) bound on the VC dimension $\gamma$. An exact bound implies that $h^2$ and $\gamma$ are close to each other.\\

\begin{figure}[hbtp]
        \centering	
                \includegraphics[scale=0.5]{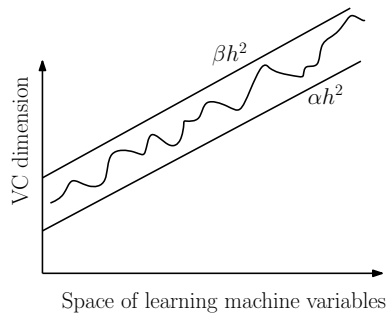}
                \caption{Illustration of the notion of an exact bound on the VC dimension. Even though the VC dimension $\gamma$ may have a complicated dependence on the variables defining the learning machine, multiples of $h^2$ bound $\gamma$ from both above and below. The exact bound $h^2$ is thus always ``close'' to the VC dimension, and minimizing $h^2$ with respect to the variables defining the learning machine allows us to find one that has a small VC dmension.}
\end{figure}\label{fig2}

Figure \ref{fig2} illustrates this notion. The VC dimension is possibly related to the free parameters of a learning machine in a very complicated manner. It is known that the number of degrees of freedom in a learning machine is related to the VC dimension, but the connection is tenuous and usually abstruse. The use of a continuous and differentiable exact bound on the VC dimension allows us to find a learning machine with small VC dimension; this may be achieved by minimizing $h$ over the space of variables defining the separating hyperplane.\\

The MCM classifier solves an optimization problem, that tries to minimize the machine capacity, while classifying all training points of the linearly separable dataset correctly. This problem is given by
\begin{equation}\label{minh1}
\operatorname*{Minimize  }_{u, v} \; h ~=~ \frac{\operatorname*{Max}_{i = 1, 2, ..., M} \; y_i(u^T x^i + v)}{\operatorname*{Min}_{i = 1, 2, ..., M} \; y_i(u^T x^i + v)},
\end{equation}
that attempts to minimize $h$ instead of $h^2$, the square function $(\cdot)^2$ being a monotonically increasing one.

In \citep{mcmneucom, mcmarxiv}, we further show that the optimization problem (\ref{minh1}) may be reduced to the problem
\begin{gather}
\operatorname*{Min}_{w, b, h} ~~h \label{objm4}\\
h \geq y_i \cdot [{w^T x^i + b}], ~i = 1, 2, ..., M \label{consm41}\\
y_i \cdot [{w^T x^i + b}] \geq 1, ~i = 1, 2, ..., M, \label{consm42}
\end{gather}
where $w \in \Re^n$, and $b, h \in \Re$. We refer to the problem (\ref{objm4}) - (\ref{consm42}) as the hard margin Linear Minimum Complexity Machine (Linear MCM). \\

Once $w$ and $b$ have been determined by solving (\ref{objm4})-(\ref{consm42}), the class of a test sample $x$ may be determined from the sign of the discriminant function
\begin{equation}\label{testresult}
 f(x) = w^T x + b
\end{equation}

The soft margin MCM allows for a tradeoff between the machine capacity and the classification error, and is gven by
\begin{gather}
\operatorname*{Min}_{w, b, h} ~~h + C \sum_{i=1}^M q_i \label{obj5}\\
h \geq y_i \cdot [{w^T x^i + b}] + q_i, ~i = 1, 2, ..., M \label{cons51}\\
y_i \cdot [{w^T x^i + b}] + q_i \geq 1, ~i = 1, 2, ..., M \label{cons52}\\
q_i \geq 0 \label{cons53}
\end{gather}

The problem (\ref{obj5})-(\ref{cons53}) tries to find a classifier with as small a VC dimension as possible, and that makes few errors on the training set. Here, $C$ determines the emphasis given to the classification error.

\begin{figure}[hbtp]
        \centering	
                \includegraphics[scale=0.5]{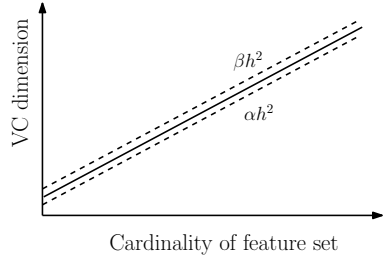}
                \caption{In the case of a linear hyperplane classifier, the VC dimension is upper bounded by $n + 1$, where $n$ is the number of features. Hence, $h^2$ is also an exact bound on the number of features used, and minimizing $h^2$ with respect to $w$ and $b$ yields a classifier that uses a small subset of discriminative features.}
\end{figure}\label{fig3}

Note that for a linear hyperplane classifier that learns from $n$-dimensional input samples, the VC dimension is upper bounded by $n+1$, as illustrated in Fig. \ref{fig3}. The figure also shows that the lower and upper bounds on the VC dimension, $\alpha h^2$ and $\beta h^2$, respectively, also are lower and upper bounds for the cardinality of the feature set. Hence, minimizing $h^2$ also yields a classifier that uses a small number of features. Note that the second term of the objective function in (\ref{obj5}) measures error on the training samples. Hence, the set of features used need to be discriminative enough on the training samples. Hence, the approach does not just choose a small \textit{list} of features, but chooses a small \textit{set} of features that has the ability to discriminate well.

\section{Experimental results}\label{experimental}
The implementations used for ReliefF \cite{relieff}and FCBF \cite{fcbf, fcbf2} are taken from \cite{zhao2010advancing}. The MCM was implemented by solving (\ref{obj5})-(\ref{cons53}) in MATLAB.

The experiments involved two stages. In the first stage, the most relevant features were extracted using the three feature selection methods that we have considered, namely, the MCM, ReliefF \cite{relieff}and FCBF \cite{fcbf, fcbf2}. All three methods take the feature matrix as input; the MCM and FCBF implementations return the indices of the features that are non-redundant. In the case of the ReliefF implementation, the routine provides the list of features along with their information gain values. We chose the highest ranked subset of features, which together constituted an arbitrarily set threshold of 40\% of the total information gain.\\

A kernel SVM with a Gaussian kernel was used to learn a classifier on the selected features in each case. The hyper-parameters used for the SVMs were chosen by a grid search. A five-fold cross validation methodology was used to assess the generalization performance of the classifiers thus learnt.

Table \ref{table1} compares the test set accuracies and the number of selected attributes for the three methods in question. The test set accuracies were computed by using a five fold cross validation methodology, and are indicated in the format (mean $\pm$ standard deviation), computed across the five folds. The test sets in most high dimensional benchmark datasets are small, owing to the small number of samples. The results in the table highlight the observation that the MCM is able to capture a small yet salient subset of discriminative features that can accurately predict the result on test set samples. This robustness may be attributed to the fact that the set of discriminative features has been selected by finding a minimum VC dimension classifier.

We have used a SVM with a RBF kernel function as the classifier to highlight the feature selection property of the MCM. Using the kernel MCM as the classifier would allow for further improvements in the test set accuracies.

\begin{table}[htbp]
\centering
\footnotesize\setlength{\tabcolsep}{2.5pt}
\caption{Feature Selection Results}
\begin{tabular}{|c|c|c|c|c|c|c|c|}
\hline
\multicolumn{ 2}{|c|}{\textbf{Dataset}} & \multicolumn{ 3}{c|}{\textbf{Features}} & \multicolumn{ 3}{c|}{\textbf{Test Set Accuracy}} \\ \hline
\multicolumn{ 2}{|c|}{\textbf{(samples x dimension)}} & \textbf{MCM} & \textbf{ReliefF} & \textbf{FCBF} & \textbf{MCM} & \textbf{ReliefF} & \textbf{FCBF} \\ \hline
\multicolumn{ 2}{|c|}{West (49 $\times$ 7129)} & 32 & 2207 & 1802 & 79.7 $\pm$ 6.8\% & 65.3 $\pm$ 3.2\% & 59.19 $\pm$ 3.1\% \\ \hline
\multicolumn{ 2}{|c|}{Artificial (100 $\times$ 2500)} & 79 & 1155 & 1839 & 82.1 $\pm$ 6.1\% & 80.8 $\pm$ 2.9\% & 80.7 $\pm$ 2.2\% \\ \hline
\multicolumn{ 2}{|c|}{Cancer (62 $\times$ 2000)} & 48 & 509 & 1346 & 77.3 $\pm$ 6.2\% & 74.6 $\pm$ 0.9\% & 75.8 $\pm$ 4.1\% \\ \hline
\multicolumn{ 2}{|c|}{Khan (63 $\times$ 2308)} & 48 & 437 & 897 & 92.7 $\pm$ 1.5\% & 89.6 $\pm$ 5.5\% & 91.3 $\pm$ 0.5\% \\ \hline
\multicolumn{ 2}{|c|}{Gravier (168 $\times$ 2905)} & 132 & 1096 & 1573 & 83.3 $\pm$ 2.6\% & 84.5 $\pm$ 1.8\% & 82.5 $\pm$ 1.6\% \\ \hline
\multicolumn{ 2}{|c|}{Golub (72 $\times$ 7129)} & 47 & 2271 & 7129 & 95.8 $\pm$ 4.2\% & 90.3 $\pm$ 4.8\% & 95.8 $\pm$ 4.2\% \\ \hline
\multicolumn{ 2}{|c|}{Alon (62 $\times$ 2000)} & 41 & 896 & 1984 & 83.8 $\pm$ 3.3\% & 82.2 $\pm$ 7.4\% & 82.1 $\pm$ 7.8\% \\ \hline
\multicolumn{ 2}{|c|}{Christensen (198 $\times$ 1413)} & 98 & 633 & 1413 & 99.5 $\pm$ 0.7\% & 99.5 $\pm$ 0.7\% & 99.5 $\pm$ 0.7\% \\ \hline
\multicolumn{ 2}{|c|}{Shipp (77 $\times$ 7129)} & 51 & 3196 & 7129 & 96.1 $\pm$ 0.1\% & 93.5 $\pm$ 2.1\% & 93.5 $\pm$ 4.4\% \\ \hline
\multicolumn{ 2}{|c|}{Singh (102 $\times$ 12600)} & 81 & 5650 & 11619 & 91.2 $\pm$ 3.9\% & 89.2 $\pm$ 2.0\% & 92.5 $\pm$ 2.7\% \\ \hline
\end{tabular}
\label{table1}
\end{table}

The table indicates that although MCM selects a much smaller subset of attributes, still the SVM classifier learnt from those features predicts as well or better than classifiers learnt using a much larger subset of features.

\section{Conclusion}\label{conclusion}
The Minimal Complexity Machine (MCM) is a recent approach that learns a classifier by minimizing an exact bound on the VC dimension. The VC dimension, that measures machine complexity, is intimately linked to the generalizing ability of a classifier. In this paper, we use the linear MCM to determine a minimal subset of features. These are then used to build a kernel SVM classifier. Experimental results on benchmark datasets show that this approach to feature selection chooses a very small number of attributes in comparison to ReliefF and FCBF. To the best of our knowledge, feature selection through minimizing the VC dimension has not been attempted before, and the present approach provides a way to use a universally applicable measure for feature selection. It has not escaped our attention that such a measure may be used in many other ways to address feature selection and other related applications.

\section*{Acknowledgement}
The authors would like to thank Prof. Suresh Chandra of the Department of Mathematics at IIT Delhi for his valuable advice and critical appraisal of the manuscript.

\begingroup
\bibliography{mcmfs2}
\bibliographystyle{plainnat}

\raggedright
\sloppy
\endgroup

\end{document}